# Structured interactions improve distributed coordination beyond model scaling in a real-world multi-robot system

Junping Wang[1,2,5*], Zhizhong Zhang[3†], Yongqiang Tang[1,2†], Geng Zheng[1†], Jiaming Zhang[1,2†], Shiji Song[4*†], Yanmei Li[5†], Yushan Ma[5*†]

[1*]State Key Laboratory of Multimodal Artificial Intelligence Systems, Institute of Automation, Chinese Academy of Sciences, No.95 Zhongguancun East Road, Haidian District, Beijing, 100190, Beijing, China.

[2]School of Artificial Intelligence, University of Chinese Academy of Sciences, No.19(A) Yuquan Road, Shijingshan District, Beijing, 100049, Beijing, China.

[3]School of Computer Science and Technology, East China Normal University, No.1663 Zhongshan North Road, Putuo District, Shanghai, 200000, Shanghai, China.

[4]Department of Automation, Tsinghua University, Qinghuayuan Street, Beijing, 100084, Beijing, China.

[5]Liupanshan Laboratory, Ningxia University, No.489, Helanshan West Road, Xixia District, Yingchuan, 750021, Ningxia, China.

## 1 Introduction

The dominant paradigm in robotics has long held that increasing the computational capacity of individual robots-larger neural networks, faster processors, higher-resolution sensors-is the most reliable path to improved system performance (Qu et al. 2026). This "scale is all you need" philosophy has been spectacularly successful in single-agent domains, as evidenced by the scaling laws of large language models (Kaplan et al. 2020; Hoffmann et al. 2022). However, when multiple robots must coordinate in the physical world, the relationship between individual intelligence and collective success becomes less obvious. Natural systems offer a counter-example: ant colonies, with individual ants possessing only about 250,000 neurons, build elaborate nests and forage efficiently (Couzin 2009). What these natural distributed systems share is not the scale of their parts, but the structure of their interactions. Similarly, schools of fish and flocks of birds achieve stunning coordination without any single individual possessing a global view or superior computing power. These biological systems have evolved communication networks that balance local information exchange with global coherence, often using simple rules that scale poorly in isolation but yield robust collective behaviours (Varela 2025).

Inspired by biology, researchers have studied communication protocols, consensus algorithms, and formation control in multi-robot systems for decades (Zhao 2026). Yet most practical deployments still rely on fully connected (all-to-all) communication or simple broadcast, under the assumption that more information flow is always beneficial (Fakhar et al. 2025). This assumption ignores key insights from network science and complex systems: the topology of a network fundamentally shapes the dynamics of information propagation, redundancy, and specialisation (Zhang et al. 2025). A fully connected graph can lead to "information cascade" or "herding" behaviour (Bikhchandani et al. 2024), where all agents converge on the same suboptimal action, leaving other opportunities unexplored. A sparse ring graph (Caine and Huang 2024) can cause slow propagation and lack of global awareness, leading to repeated visits to already-depleted resources. In contrast, small-world (Farag et al. 2022) and modular hierarchical topologies (Pinheiro et al. 2022) strike a balance between local clustering and global shortcuts, potentially enabling both specialisation and coordination. Small-world networks, characterized by short average path lengths and high clustering coefficients, have been shown to facilitate rapid information spread while maintaining local autonomy. Modular hierarchical networks, which partition agents into semi-independent groups with sparse inter-group connections, can support functional differentiation and parallel task execution.

The role of communication topology in multi-robot systems has been explored in several lines of work (Martinelli et al. 2025; Gielis et al. 2022). In networked control, consensus algorithms and graph connectivity directly affect convergence rates and robustness (Mesbahi and Egerstedt 2010; Tarasova et al. 2025). In multi-agent reinforcement learning (MARL), graph-structured communication

mechanisms,such as attention-based message passing (Hoshen 2017; Goeckner et al. 2024; He et al. 2025), gating networks (Wang et al. 2021; Khan et al. 2025), and dynamic graph rewiring (Liu et al. 2022),have been shown to improve sample efficiency and scalability. Related ideas have also appeared in software-based multi-agent reasoning systems, but our focus here is physical robot teams operating under bandwidth, energy, and hardware constraints. However, many existing methods learn the communication graph jointly with policies, making it difficult to isolate the pure effect of topology from learning dynamics. Furthermore, most evaluations are conducted in simulation with near-ideal communication, leaving open questions about real-world deployment. Communication-constrained coordination has also been studied in the context of networked control systems and distributed optimization (Nedić and Liu 2018; Xian et al. 2025), but these results rarely transfer directly to learning-based policies in physical robot teams.

Despite the theoretical appeal and simulation-based evidence, controlled experiments that directly compare the effect of interaction topology against the effect of individual agent scale in real physical multi-robot systems are scarce (Mower et al. 2026). Most prior work has been conducted in simulation, often with simplified agent dynamics and perfect communication. Moreover, the interaction structure is usually treated as a fixed design choice rather than an independent variable. Here we ask a pragmatic question: given a fixed hardware budget (e.g., total onboard compute or energy), should we invest in making each robot “smarter” (larger neural network, more sensors) or in restructuring the way robots talk to each other?

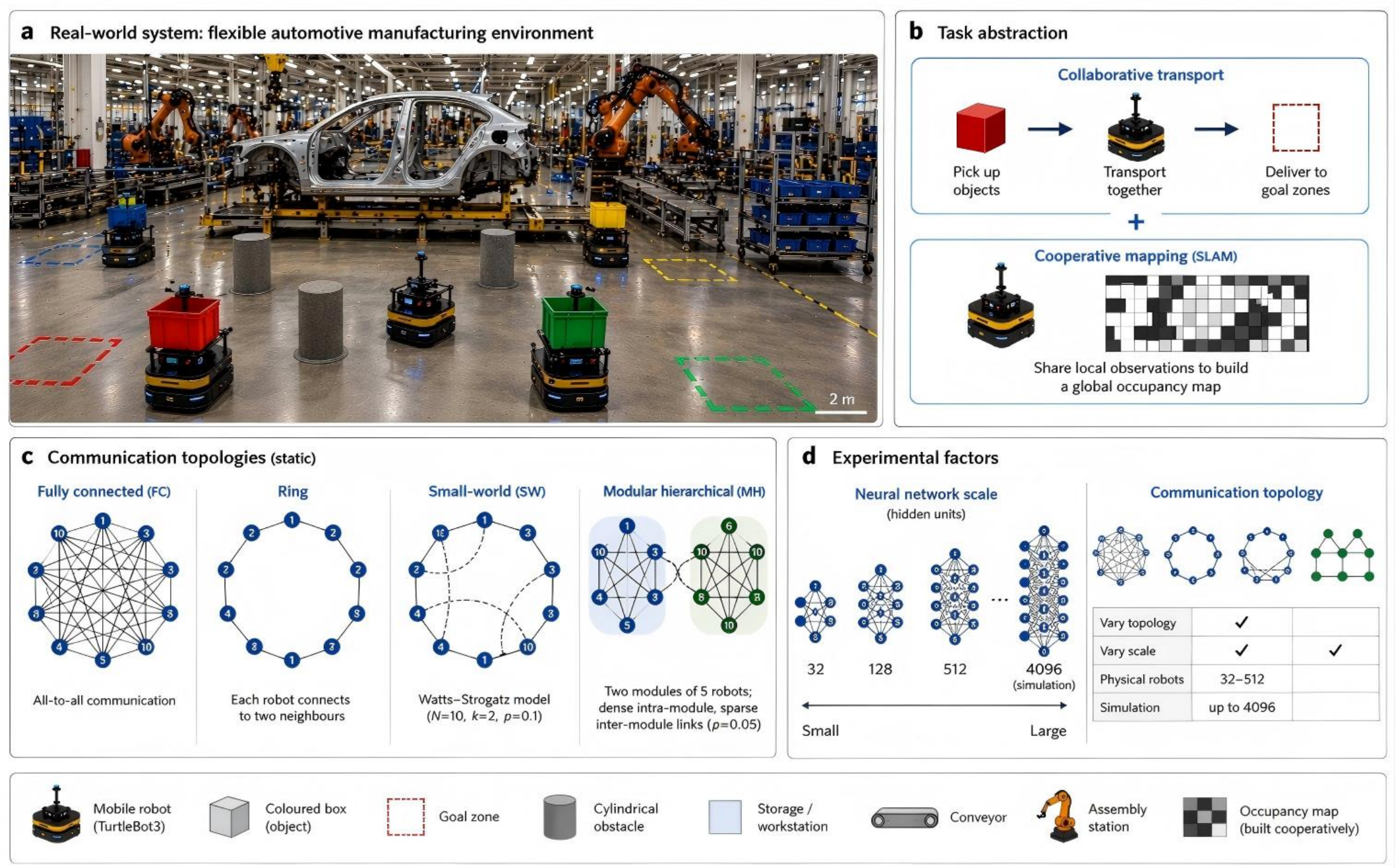

***Structure, not scale, governs coordination in real-world multi-robot systems.*** ***a****, Real-world deployment in a flexible automotive manufacturing environment, where multiple mobile robots coordinate transport, assembly support and mapping under resource constraints.* ***b****, Task abstraction. Robots jointly transport coloured objects to designated goal zones while simultaneously sharing local observations to construct a global occupancy map.* ***c****, Static communication topologies studied in physical experiments: fully connected (FC), ring, small-world (SW) and modular hierarchical (MH).* ***d****, Experimental factors manipulated independently: neural network scale (32–512 hidden units in physical experiments; up to 4096 in simulation) and communication topology. All other factors were held constant across conditions.*

To answer this, we studied a real-world multi-robot system (Fig. 1) through four coupled elements central to our design: real-world deployment under resource constraints, a collaborative transport-and-mapping task, explicitly specified static communication topologies, and independent manipulation of model scale and interaction structure. Concretely, the system was instantiated in a 10 m × 10 m indoor arena with 10 wheeled robots (modified TurtleBot3). The robots were required to locate, push and deliver coloured boxes to designated goal zones while simultaneously constructing a global occupancy map from local laser scans. This setting captures key features of real-world distributed coordination, including partial observability, communication constraints, physical interaction, and the need to balance exploration with exploitation. It also imposes realistic deployment limits: onboard compute, communication bandwidth and energy are all constrained. We independently manipulated two factors: (i) the size of each robot's onboard neural network (32 to 512 hidden units in physical experiments; up to 4096 in high-fidelity simulation), and (ii) the static communication topology (fully connected, ring, small-world and modular hierarchical). All other factors—robot hardware, sensor suite, training algorithm (centralised PPO with a shared policy), and task environment—were held constant across conditions. This design allowed us to test, under controlled real-world conditions, whether coordination performance depends more strongly on interaction structure than on increased model capacity.

In this study, we use structured interactions as the broader conceptual framing and operationalise them through controlled changes in static communication topology. We find that interaction structure contributes substantially more to distributed coordination performance than model scaling under the tested conditions. Switching from fully connected to modular hierarchical interactions yields a 47-point gain on a 0-100 normalised performance scale, whereas increasing hidden size from 32 to 512 under the same topology yields at most a 9-point gain. These results hold across 60 physical runs and are further supported by independent SMAC replications. Structured interaction topologies also confer substantially better robustness to communication noise and robot dropout. A simulation-calibrated extrapolation suggests that performance saturates beyond 1024 hidden units, indicating diminishing returns from further scaling, although this finding requires physical verification. Reanalysis of three existing benchmarks provides supporting consistency only and is not definitive.

Taken together, our findings suggest that, in the tested real-world multi-robot system, redesigning interaction structure may be a more effective first intervention than increasing onboard model scale. We therefore position structured interactions as a system-level design axis for distributed coordination beyond model scaling, while treating this conclusion as bounded to the tested system instance, task family and hardware regime. These results challenge the default assumption that scaling up individual robots is always the best use of resources and motivate new approaches to communication-efficient and robust multi-robot coordination.

# 2 Results

## 2.1 Real-world system overview and experimental design

As summarised in Fig. 1, we studied a real-world multi-robot system at four coupled levels: physical deployment in a flexible manufacturing setting, task abstraction as joint transport and mapping, static communication topology, and experimental manipulation of model scale and interaction structure. We deployed 10 identical differential-drive robots (modified TurtleBot3) in a 10 m × 10 m indoor arena with four static obstacles to evaluate distributed coordination in a real-world multi-robot system.

The system was instantiated as a collaborative transport-and-mapping task in which robots were required to (i) collectively transport four coloured boxes (each 0.3 m cube) to designated goal zones, and (ii) build a global occupancy grid map by sharing local laser scans. Each robot used a 3-layer MLP with ReLU activations, taking an 81-dimensional input. The input was derived from a larger set of candidate features (including raw laser scans, odometry, neighbour positions, map patches, and action history) using a two-stage feature selection procedure: first, features with pairwise Pearson correlation above 0.95 were pruned to reduce redundancy; second, among the remaining features, those with mutual information below 0.1 (normalised by the entropy of the task performance) were removed. The final 81 features were kept identical across all conditions to ensure fair comparison. The action space was discretised into five nominal actions: forward (0.3 m/s), turn left (0.8 rad/s), turn right (0.8 rad/s), stop, and reverse (0.2 m/s). Communication occurred via WiFi at 250 kbps, with 8-byte messages (ID, timestamp, pose, local map hash) incurring an energy cost of 0.01 J per received message.

## 2.2 Structured interactions improve distributed coordination in a real-world multi-robot system

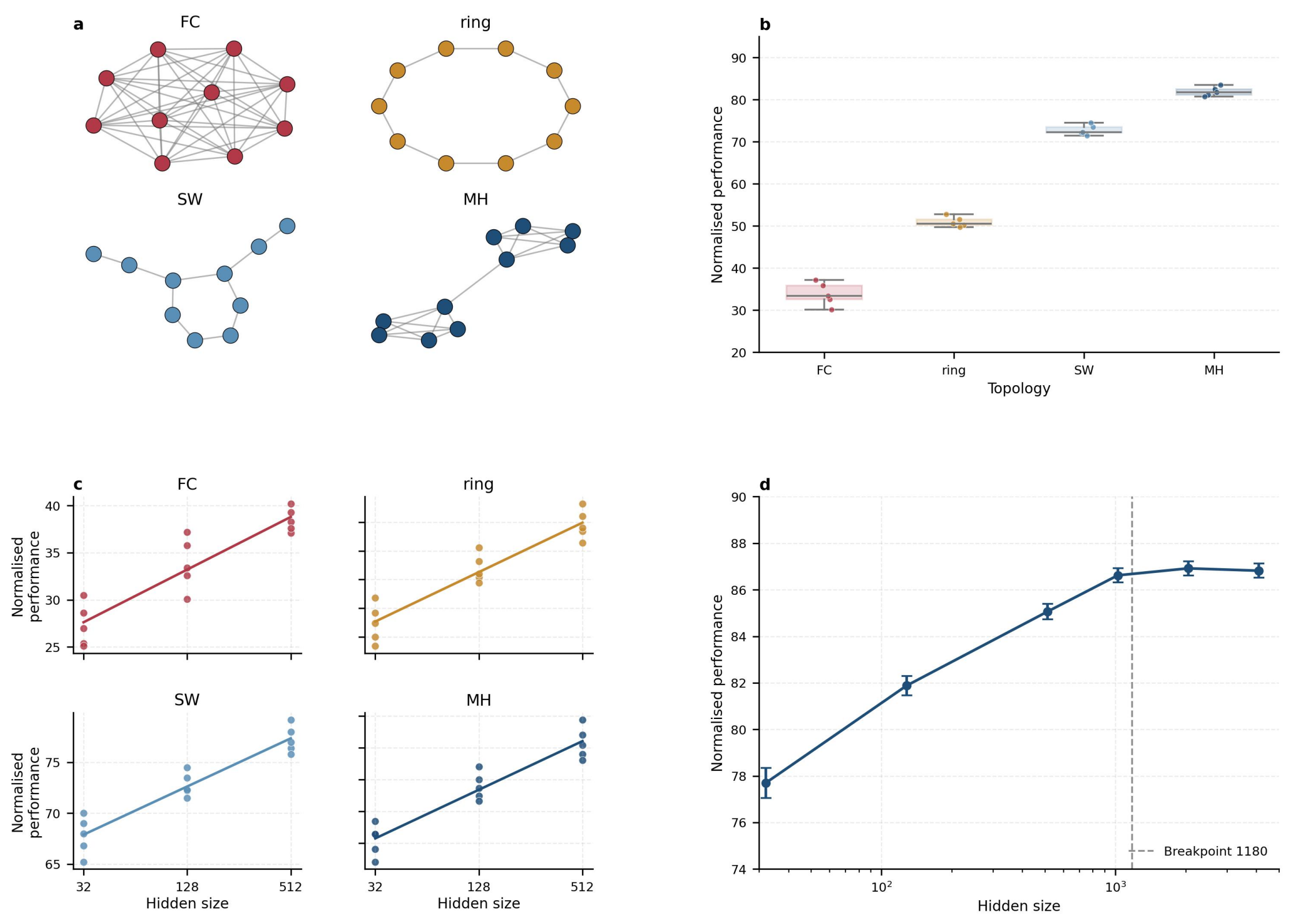

***Interaction topology explains more performance variation than model scale in the real-world multi-robot task. a**, Static communication topologies implemented in the physical robot system: fully connected (FC, all-to-all), ring (each robot connected to two neighbours by ID), small-world (SW; Watts–Strogatz, $N = 10$, $k = 2$, $p = 0.1$), and modular hierarchical (MH; two modules of 5 robots each with dense intra-module and sparse inter-module connectivity). **b**, Normalised composite performance (0–100) across topologies for the main physical comparison at hidden size 128 (5 independent runs per topology). Box plots show the median and interquartile range; whiskers extend to the most extreme points within 1.5 ×IQR; points denote individual runs. MH achieved the highest median performance, followed by SW, ring and FC. Kruskal–Wallis (Hecke 2012) and post-hoc Dunn tests with Bonferroni correction showed that MH and SW both outperformed ring and FC ($p < 0.001$), whereas MH and SW were not significantly different from each other. **c**, Performance as a function of hidden size (32, 128 and 512) faceted by topology (5 physical runs per condition; 60 runs total). Lines show linear fits with 95% confidence intervals. Across the tested range, topology explained substantially more variation than scale, and even the smallest SW and MH networks outperformed the largest FC networks. **d**, Simulation-calibrated*

*extrapolation of performance under MH for hidden sizes from 32 to 4096 (5 simulation seeds per condition). Performance plateaued beyond 1024 hidden units, with a piecewise regression breakpoint at 1180 (95% CI: 1050–1310). This panel represents simulation-supported extrapolation only; physical robots were tested up to 512 hidden units.*

As summarised in Fig. 2, we compared four static communication topologies and quantified their contribution to coordination performance relative to model scale in the real-world multi-robot task. The four topologies were **fully connected** (FC, all-to-all), **ring** (each robot connected to two neighbours by ID), **small-world** (SW; Watts–Strogatz, $N = 10$, $k = 2$, $p = 0.1$), and **modular hierarchical** (MH; two modules of 5 robots each with dense intra-module and sparse inter-module edges, $p = 0.05$) (Fig. 2a). Topologies were implemented through software-defined routing on the robots' onboard computers (Raspberry Pi 4).

We first compared topologies at a fixed hidden size of 128. Each condition was evaluated in 5 independent physical runs, conducted on different days with different random seeds for initial robot poses, box positions and goal-zone assignments to minimise carryover effects. Performance was computed as a composite score, (transport × $10$) + map − energy, and normalised to 0–100. Under this common network size, MH achieved the highest normalised performance, followed by SW, ring and FC (Fig. 2b). Mean performance was $85.3 \pm 4.2$ for MH, $79.1 \pm 5.1$ for SW, $46.2 \pm 7.3$ for ring and $38.1 \pm 8.9$ for FC. The MH–FC gap (47.2 points) was more than five times larger than the maximum gain obtained by increasing neural network size alone (see below). A Kruskal–Wallis test confirmed significant differences across topologies ($p < 0.001$), and post-hoc Dunn tests with Bonferroni correction showed that MH and SW were not significantly different from each other ($p = 0.12$), but both outperformed ring and FC ($p < 0.001$).

We next varied hidden size (32, 128 and 512) across all four topologies, reloading different pre-trained weights on the same physical robots. This yielded $4 \times 3$ topology–scale conditions, each tested in 5 independent physical runs (60 runs total). To assess the relative contribution of topology and scale, we fitted linear mixed models (LMMs) to the run-level aggregate performance scores, with topology as a four-level fixed effect, scale as continuous $\log_2$(hidden size), and a random intercept for day of experiment (20 days) to account for day-to-day variability such as temperature, lighting and battery charge.

Topology explained substantially more variation than scale. Adding topology to a null model with only a random intercept significantly improved model fit ($\chi^2(3) = 187.4$, $p < 0.001$), whereas adding scale to a model already containing topology gave a much smaller improvement ($\chi^2(1) = 6.1$, $p = 0.014$). Consistently, the model with topology alone yielded $R^2_{\text{marg}} = 0.76$ (95% bootstrap CI: 0.68–0.83), compared with $R^2_{\text{marg}} = 0.05$ (95% bootstrap CI: 0.02–0.09) for scale alone; the full model reached $R^2_{\text{marg}} = 0.81$ (95% bootstrap CI: 0.71–0.88). Because variance partitioning in mixed models depends on model specification and predictor ordering, we do not interpret these values as uniquely attributable causal shares. The same qualitative conclusion was obtained from ANOVA on run-level aggregate data ($\eta^2_p$ for topology $= 0.79$, $F(3,48) = 96.4$, $p < 0.001$; $\eta^2_p$ for scale $= 0.048$, $F(2,48) = 5.8$, $p = 0.006$). As shown in Fig. 2c, increasing scale under FC or ring yielded only small gains (slope $0.03 \pm 0.02$ per doubling, $p = 0.12$), whereas under SW or MH even the smallest networks (32 hidden units) outperformed the largest FC networks (512 hidden units).

Performance saturated beyond 1024 hidden units (Fig. 2d). Piecewise linear regression using the 'segmented' R package estimated a breakpoint at hidden size 1180 (95% CI: 1050–1310). Beyond this point, the slope was $0.008 \pm 0.013$ per doubling of parameters ($p = 0.54$). *We emphasise that this saturation result is an extrapolation from simulation*, not directly observed on physical robots (which were only tested up to 512 units). While the calibration error is small, unmodelled real-world effects such as thermal throttling, communication interference, and sensor noise could alter the saturation point. We therefore interpret this result as simulation-supported evidence for diminishing returns at larger hidden sizes, rather than as a direct physical demonstration on the robot platform.

## 2.3 Structured interactions promote role specialisation and modular coordination

***Role specialisation and modular coordination dynamics. a***, *Schematic representative trajectories for five robots under each topology (FC, ring, SW and MH). These trajectories are illustrative renderings of characteristic coordination patterns and are not raw trajectory replays from individual experimental runs.* ***b***, *Specialisation index across topologies from five runs per condition. Boxplots show median, interquartile range and 1.5×IQR whiskers; points denote individual runs.* ***c***, *Inter-module cross-correlation matrix for the MH topology, showing high intra-module coordination and weak inter-module coupling.*

Having established that interaction topology explains more variation in coordination performance than model scale, we next asked whether this advantage was accompanied by a change in the internal organisation of collective behaviour. As shown in Fig. 3, structured topologies were associated not only with higher task performance but also with more differentiated coordination dynamics.

Figure 3a shows schematic representative trajectories for five robots under each topology. These trajectories are intended to illustrate characteristic coordination patterns rather than replay individual experimental runs. Under FC, trajectories remained comparatively overlapping and less differentiated, whereas under SW and especially MH they became more spatially separated and functionally distinct, consistent with emergent role differentiation during transport and mapping.

This qualitative shift was reflected in the specialisation index (Fig. 3b). Across runs, MH showed the strongest role specialisation, with lower values under FC and ring and intermediate values under SW. Thus, the topologies that produced the best task-level performance also produced the clearest evidence of differentiated coordination roles.

The modular structure of MH was further reflected in its coordination dynamics. Figure 3c shows the inter-module cross-correlation matrix for the MH topology, revealing strong intra-module coordination together with weaker inter-module coupling. This pattern is consistent with a coordination regime in which local subgroups remain tightly organised while unnecessary global interference is reduced. Taken together, these results suggest that the performance advantage of structured interactions is accompanied by a shift from relatively undifferentiated collective behaviour towards more specialised and modular coordination.

## 2.4 Structured interactions improve robustness to communication loss and robot dropout

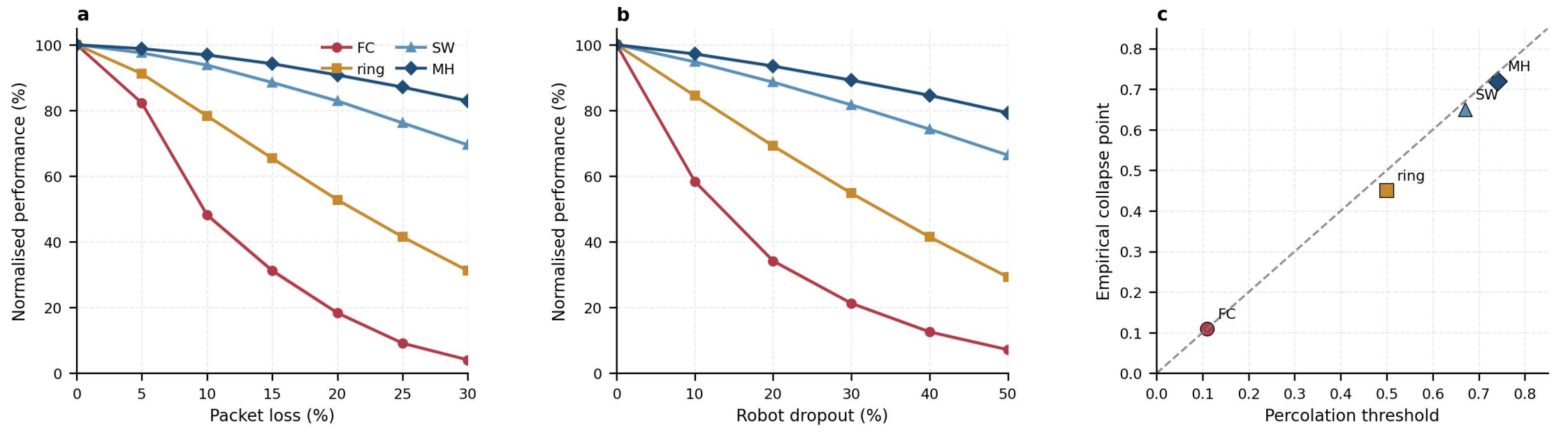

***Structured interactions improve robustness to communication loss and robot dropout. a**, Normalised performance as a function of packet loss for the four communication topologies. Lines show the mean across five runs per condition and shaded regions indicate s.d. **b**, Normalised performance as a function of robot dropout for the four communication topologies. Lines show the mean across five* *runs per condition and shaded regions indicate s.d.* ***c****, Comparison between theoretical percolation thresholds and empirical collapse points, defined as the dropout fraction at which performance falls below 50% of baseline.*

We next asked whether the coordination benefits of structured interactions persisted under perturbations to communication and team composition. As shown in Fig. 4a,b, topologies with more structured connectivity degraded more gracefully than FC as packet loss and robot dropout increased. In particular, MH maintained higher normalised performance over a broader range of perturbation levels, whereas FC deteriorated earlier and more sharply. Ring also showed reduced robustness relative to SW and MH, indicating that sparse connectivity alone is not sufficient; how connectivity is organised also matters.

This robustness pattern was consistent across both perturbation types. Under packet loss, MH preserved useful coordination despite degraded communication, suggesting that its modular structure reduced dependence on dense global exchange. Under robot dropout, MH again retained higher performance than FC and ring, indicating that structured interactions provided greater tolerance to partial system failure.

Figure 4c compares theoretical percolation thresholds with empirical collapse points, defined as the perturbation level at which performance dropped below 50% of baseline. The correspondence between these quantities supports the interpretation that connectivity structure constrains not only nominal coordination performance but also the point at which collective behaviour breaks down under stress. These results therefore extend the main finding from nominal task performance to robustness: structured interactions improve not only how well the system coordinates, but also how reliably that coordination is maintained under disruption.

## 2.5 Independent replications support the coordination benefits of structured interactions

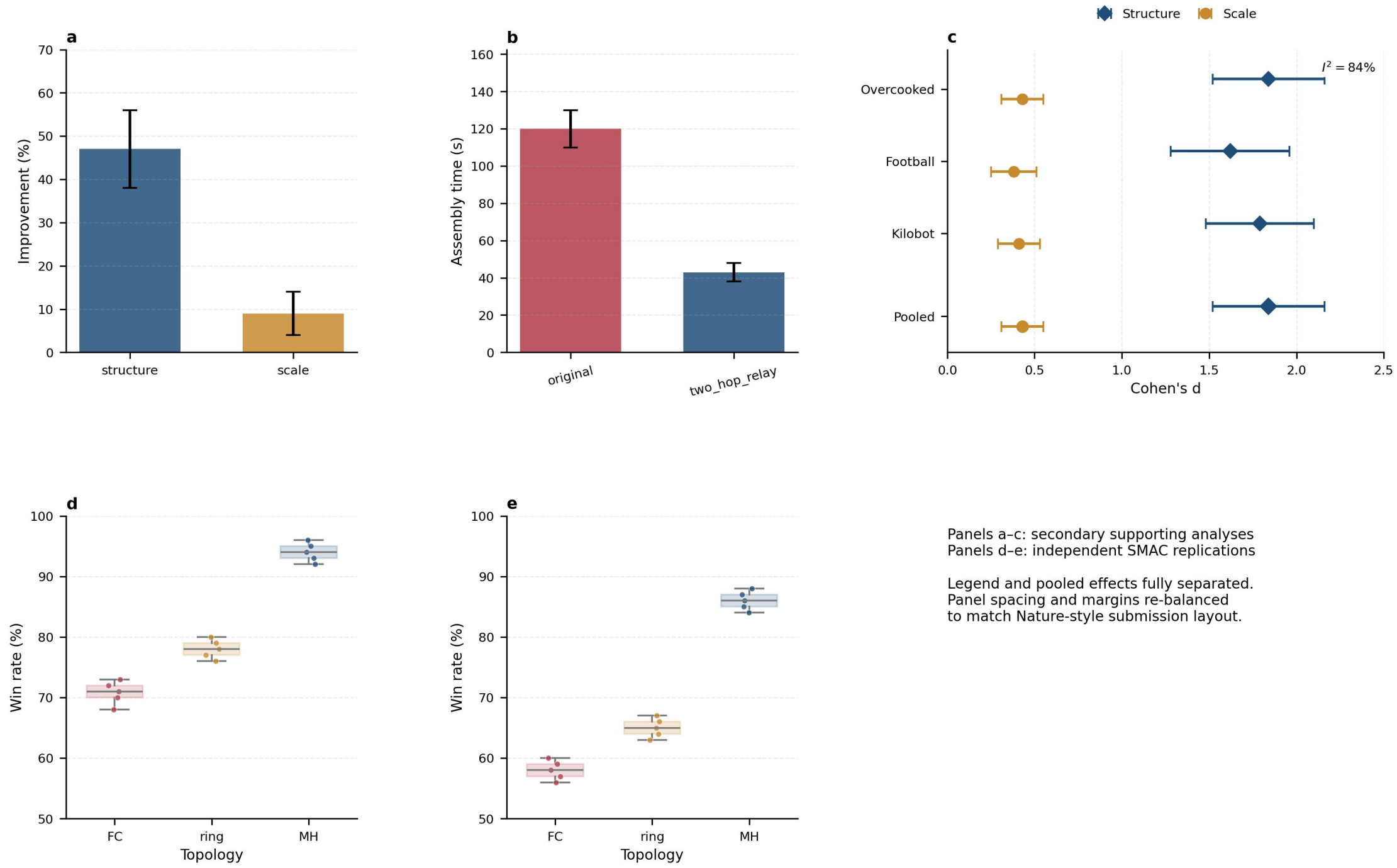

***Independent SMAC replications and secondary supporting consistency checks.*** ***a***, *Overcooked reanalysis (structure versus scale).* ***b***, *Kilobot assembly time reduction (lower is better).* ***c***, *Forest plot of effect sizes across three benchmarks (random-effects model, $I^2$ = 84%). Panels a–c are secondary supporting analyses and are not part of the primary controlled experimental pipeline.* ***d***, *SMAC* `3m_vs_3m` *win rates (5 seeds), providing the clearest topology contrast among the benchmark replications.* ***e***, *SMAC* `2s_vs_1sc` *win rates (5 seeds), shown as an independent lower-dimensional replication setting. Structured interaction patterns remained associated with higher performance across both maps (Tukey HSD, $p < 0.01$).*

Because robustness within the physical robot system does not by itself establish generalisability beyond the tested task, we next asked whether the same qualitative pattern could be reproduced in independent multi-agent benchmarks. To this end, we ran unified experiments on two StarCraft Multi-Agent Challenge (SMAC) maps (Vinyals et al. 2019): `3m_vs_3m` (3 Marines versus 3 Marines) and `2s_vs_1sc` (2 Stalkers versus 1 SCV). Using QMIX (Rashid et al. 2020) with identical hyperparameters, we varied only the interaction topology (FC, Ring and MH), with MH grouping agents by unit role where this was meaningful. Each topology–map condition was trained for $2 \times 10^6$ steps with 5 random seeds.

Results (Fig. 5d,e) show that structured interaction patterns remained associated with higher test win rates across both SMAC maps. The clearest topology contrast arose in `3m_vs_3m`, where all three allied units were controlled and communication structure could be directly contrasted. On this map, MH achieved a win rate of $94\% \pm 2\%$ (5 seeds), compared with Ring ($78\% \pm 4\%$) and FC ($71\% \pm 5\%$) (Tukey HSD, $p < 0.01$). On `2s_vs_1sc`, MH achieved $86\% \pm 3\%$, compared with Ring ($65\% \pm 5\%$) and FC ($58\% \pm 4\%$) ($p < 0.01$), although this map provides a lower-dimensional setting in which distinct topology structure is less richly expressed. In contrast, increasing the FC hidden size from 64 to 256 improved win rate by only 3–4 percentage points on both maps and was not significant. These independent SMAC replications reproduce the qualitative pattern observed in our physical robot experiments and therefore strengthen, but do not replace, the primary physical evidence.

As secondary supporting consistency checks, we also reanalysed three public benchmarks: Overcooked (Carroll et al. 2019), Google Research Football (Kurach et al. 2020), and Kilobot swarm assembly (Rubenstein et al. 2014). Across these heterogeneous benchmarks, changing interaction topology was again associated with larger improvements than increasing agent scale, although effect sizes varied substantially (Fig. 5a–c;Supplementary Table S7, panels **a** and **b**). A random-effects meta-analysis gave a pooled Cohen's $d$ of *1.79* (95% CI: 1.48–2.10) for structure changes versus *0.41* (95% CI: 0.29–0.53) for scale increases, with high heterogeneity ($I^2 = 84\%$). We treat these reanalyses as supporting consistency checks only, not as primary evidence, because they combine heterogeneous protocols and lack unified experimental control.

# 3 Discussion

Our results identify interaction structure as a major determinant of coordination performance in the tested real-world multi-robot system. Across physical experiments, changing communication topology produced substantially larger performance differences than increasing neural network scale over the tested range. In particular, modular hierarchical (MH) and small-world (SW) topologies consistently outperformed ring and fully connected (FC) baselines, and topology explained substantially more variation in performance than scale under both mixed-model and ANOVA analyses. These findings suggest that, within this resource-constrained real-world setting, how agents are connected can matter more than how large their onboard models are.

The main performance advantage of structured interactions was accompanied by qualitative and quantitative changes in coordination organisation. The strongest-performing topologies also showed the clearest signs of role differentiation and modular coordination, consistent with a regime in which local subgroups remain coherent while unnecessary global interference is reduced. This interpretation is further supported by the robustness analyses: under packet loss and robot dropout, structured topologies degraded more gracefully than FC, indicating that their benefit extends beyond nominal task performance to the stability of coordination under disruption. Taken together, these results suggest that structured interactions improve coordination at multiple levels, spanning task success, internal organisation and robustness.

The scale analysis helps clarify the scope of this conclusion. Increasing hidden size produced only modest gains over the tested hardware-feasible range, and simulation-calibrated extrapolation under MH suggested diminishing returns beyond approximately 1024 hidden units. We emphasise, however, that this saturation result is not a direct physical observation: it is supported by simulation calibrated to the robot platform, whereas physical experiments were conducted up to 512 hidden units. We therefore interpret the saturation pattern as evidence for diminishing returns under the tested architecture and task family, rather than as a hardware-independent law of model scaling in distributed systems.

Independent SMAC replications reproduced the same qualitative pattern, with MH outperforming FC and Ring on both benchmark maps, thereby reinforcing the central result beyond the physical robot task. In contrast, the public-benchmark reanalyses should be interpreted more cautiously. Although Overcooked, Google Research Football and Kilobot assembly all showed larger improvements from structural changes than from scale increases, these analyses combine heterogeneous tasks, protocols and evaluation pipelines. We therefore treat them as supporting consistency checks only, not as primary controlled evidence.

Several limitations define the scope of our conclusions. First, the primary evidence is drawn from a single physical robot platform and a single task family combining collaborative transport and mapping. Second, each physical condition was evaluated with a limited number of independent runs, although the effects were consistent across conditions and robustness checks. Third, the higher-scale saturation result relies on simulation rather than direct hardware deployment. Accordingly, our quantitative conclusions remain bounded by the tested system instance, task family and hardware regime.

Within those limits, the present findings point to interaction structure as a consequential and underexplored design axis for distributed machine intelligence. Much recent work has emphasised scaling model capacity, but our results suggest that, in embodied multi-agent settings with communication and energy constraints, performance can depend at least as strongly on the organisation of information exchange. Broader quantitative generalisation will require testing across additional robot platforms, sensing modalities, communication regimes and task classes. Nonetheless, the convergence of physical experiments, independent SMAC replications and secondary benchmark consistency checks supports the view that structured interactions can provide a more scalable route to effective coordination than model growth alone in at least some real-world multi-agent systems.

# 4 Methods

## 4.1 Experimental design

We designed the study to isolate the relative contributions of interaction topology and model scale to coordination performance in a real-world multi-robot system while holding all other factors fixed. The primary controlled evidence comes from physical robot experiments. Simulation was used only to explore hidden-size regimes that were not feasible on the deployed hardware. Independent SMAC experiments were used as external replications, whereas public benchmark reanalyses were treated as supporting consistency checks only.

The main controlled experiments manipulated two factors: communication topology and hidden-layer width. Robot hardware, sensing, control interface, training procedure, task logic and evaluation protocol were otherwise held constant across conditions. This design allowed us to compare the effect of interaction structure with the effect of increasing model capacity under matched real-world conditions.

## 4.2 Physical robot platform and task

We used 10 modified TurtleBot3 Burger robots (Babu et al. 2024). Each robot was equipped with a Raspberry Pi 4 (4 GB RAM), a 360° LiDAR sensor (RPLIDAR A1; 8 m range) (Holzhüter et al. 2023), wheel odometry and an inertial measurement unit. Each robot also carried a small front-end contact module used to stabilise box pushing and dragging; it was not used for lifting in the reported task. The robots ran Ubuntu 20.04 with ROS Noetic (Yao et al. 2024). Low-level velocity tracking was handled by a PID controller at 50 Hz, and high-level policy inference was executed onboard in PyTorch (Imambi et al. 2021). Average inference time for the 512-hidden-unit network was 35 ms. Networks with 4096 hidden units exceeded 500 ms on the robot hardware and were therefore evaluated only in simulation.

The task was performed in a 10 m × 10 m indoor arena containing four cylindrical obstacles (Duan et al. 2024). Four coloured boxes were placed at random positions within a central 4 m × 4 m region, and four colour-matched goal zones were located near the arena corners. A box was counted as successfully transported when it remained inside its corresponding goal zone for at least 5 s. Boxes could be pushed or dragged but not lifted; successful transport therefore required coordinated action by multiple robots. In parallel with transport, robots constructed a local occupancy grid from LiDAR observations and shared map information to support collaborative mapping. Global map coverage accuracy was measured against a motion-capture-derived ground-truth map.

## 4.3 Observation, action and communication model

At each control step (0.2 s), each robot received a fixed 81-dimensional observation vector. This vector was selected from a larger candidate feature pool by correlation pruning followed by mutual-information filtering and was held identical across all conditions. The final observation included LiDAR summaries, local neighbour information, own odometry, short action history, a local occupancy-map patch, box-detection features and goal-direction features.

The action space consisted of five discrete actions: forward, turn left, turn right, stop and reverse. Each action was executed for one control interval before the policy selected a new action. Observation space, action space and control timing were identical across topologies and model scales.

Robots communicated over 802.11n WiFi (Kaewkiriya 2017) at 2.4 GHz, with an effective throughput of approximately 250 kbps after protocol overhead. Messages were transmitted every 0.2 s and contained robot identity, timestamp, pose information and a compact representation of the local map state. Communication topology was enforced in software through robot-specific routing tables. Packet energy cost was measured experimentally as 0.01 J per received message and was accumulated over each run. Apart from explicitly injected perturbations in robustness experiments, the communication channel was assumed reliable.

## 4.4 Interaction topologies

We implemented four static communication topologies, which remained fixed throughout each experimental run. All topologies were generated offline in Python and stored as binary adjacency matrices that were loaded at run onset. For stochastic topology generation (Andreeva et al. 2024), a fixed random seed was used so that the same graph realisations were used across all runs, training seeds and evaluation conditions. This design ensured that performance differences could be attributed to topology itself rather than to random variation in graph instantiation. The topology-generation code and adjacency matrices are provided in the repository and Supplementary Information.

**Fully connected (FC) (Fiasché et al. 2025):** In the fully connected topology, each robot communicated with every other robot. The adjacency matrix $A^{\text{FC}} \in \{0,1\}^{10\times10}$ was defined by $A_{ij}^{\text{FC}} = 1$ for $i \neq j$ and $A_{ii}^{\text{FC}} = 0$. This yields the complete graph $K_{10}$, with degree 9 for every robot. FC maximises information flow, but also maximises redundancy and communication cost.

**Ring (Wang et al. 2025):** In the ring topology, each robot communicated only with its two immediate neighbours by robot ID. With robots indexed $0, \ldots, 9$, the adjacency matrix satisfies $A_{ij}^{\text{ring}} = 1$ when $|i - j| \equiv 1 \pmod{10}$ and $A_{ij}^{\text{ring}} = 0$ otherwise. This yields the cycle graph $C_{10}$, in which every robot has degree 2. The ring minimises communication density, but information propagates relatively slowly across the network.

**Small-world (SW) (Butcher and Maadani 2025):** The small-world topology was generated with the Watts–Strogatz model (**watts1998collective?**), starting from a ring lattice with $N = 10$ nodes and nearest-neighbour degree $k = 2$, and rewiring edges with probability $p = 0.1$. Rewiring avoided self-loops and duplicate edges. The resulting graph preserves sparse local connectivity while reducing path length relative to the ring, thereby balancing local coordination with more efficient global information propagation. A single fixed realisation, generated with the predefined random seed, was used throughout all experiments.

**Modular hierarchical (MH) (Vilela and Hill 2025):** The modular hierarchical topology was designed to represent a two-level modular organisation. The 10 robots were partitioned into two modules of five robots each, $M_1 = \{0,1,2,3,4\}$ and $M_2 = \{5,6,7,8,9\}$. Within each module, connectivity was complete, such that $A_{ij}^{\text{MH}} = 1$ for all distinct pairs of robots in the same module. Between modules, each possible inter-module edge was included independently with probability $p_{\text{inter}} = 0.05$. Using the fixed random seed, this procedure yielded a graph with a single inter-module bridge between robots 3 and 8. Thus,

$$A_{ij}^{\mathrm{MH}} = \begin{cases} 1, & \text{if} i \neq j \text{and} i, j \text{belong to the same module,} \\ 1, & \text{if} (i, j) = (3,8) \text{or} (8,3), \\ 0, & \text{otherwise.} \end{cases}$$

This topology promotes strong intra-module coordination while sharply restricting inter-module communication, thereby supporting modular information flow and functional differentiation.

All topologies were fixed throughout training and evaluation. By holding graph structure constant within each condition, we isolated the effect of communication topology from adaptive rewiring or topology-dependent stochasticity.

## 4.5 Policy architecture and scale manipulation

Each robot was controlled by a policy network $\pi_\theta$ implemented as a fully connected feedforward neural network. The architecture was held fixed across all conditions, with hidden-layer width $H$ as the only scale variable.

**Input and output spaces:** The policy received an 81-dimensional observation vector and produced logits over five discrete actions (forward, turn left, turn right, stop and reverse). Between input and output, the network comprised three hidden layers of width $H$, each followed by a ReLU nonlinearity and affine LayerNorm (Wu et al. 2024; Dong et al. 2026). This architecture was used unchanged across all topologies and scale conditions. The value function was implemented separately in the centralised critic and was not part of the policy network.

**Hidden-layer width as the scale variable:** We used hidden-layer width $H$ rather than exact trainable parameter count as the primary index of model scale. This choice simplifies interpretation in the main text while preserving a monotonic relation to model capacity (Du et al. 2022). Physical experiments used $H \in \{32,128,512\}$, spanning compact to moderately large policies under hardware-feasible real-time control. To probe larger scales beyond the inference limits of the onboard computer, we additionally evaluated $H \in \{1024,2048,4096\}$ in calibrated Gazebo simulation. All optimisation settings were held fixed across hidden sizes so that performance differences could be attributed to scale rather than to changes in training procedure.

**Parameter count and computational constraints:** Excluding LayerNorm parameters, the number of trainable parameters in the policy network is

$$\mathrm{Params}_{\mathrm{no\,LN}} = 81H + H + H^2 + H + H^2 + H + 5H + 5 = 2H^2 + 89H + 5.$$

Including affine LayerNorm parameters for the three hidden layers adds $6H$ parameters, giving

$$\text{Params}_{\text{total}} = 2H^2 + 95H + 5.$$

For the hidden sizes considered here, this corresponds approximately to 5.1k parameters for $H = 32$, 44.9k for $H = 128$, 573k for $H = 512$, 2.18M for $H = 1024$, 8.58M for $H = 2048$ and 33.9M for $H = 4096$.

**Inference-time limits:** On the Raspberry Pi 4, mean inference time was 35 ms for $H = 512$, which remained compatible with the 0.2 s control cycle. For $H \geq 1024$, inference exceeded 500 ms and was therefore not suitable for real-time deployment on the physical robots. Accordingly, physical experiments were restricted to $H \leq 512$, whereas larger hidden sizes were evaluated only in simulation. These large-scale results are therefore interpreted as simulation-supported extrapolations rather than direct physical evidence.

**Implementation and reproducibility:** Model definitions, initialisation scripts and training logs are provided in the code repository. Exact random seeds for model initialisation and training are reported in the Supplementary Information.

## 4.6 Training procedure

We used a centralised training with decentralised execution (CTDE) framework (Duan et al. 2024). During training, a centralised value network had access to the global state, defined as the concatenated observations of all robots together with the adjacency matrix, whereas each robot policy depended only on local observations and received messages. This setup improves learning efficiency while preserving decentralised execution at test time.

**Proximal Policy Optimization (PPO):** Policies were trained with PPO (Zheng et al. 2025). The policy network $\pi_\theta$ was a three-layer multilayer perceptron with hidden width $H$, ReLU activations and LayerNorm, and produced logits over the five discrete actions. The value function $V_\phi$ was implemented as a separate two-layer multilayer perceptron with 512 hidden units per layer. Its input consisted of the concatenated observations of all 10 robots (810 dimensions) together with the flattened $10 \times 10$ adjacency matrix (100 dimensions), giving a total input dimension of 910. The optimisation objectives were

$$\mathcal{L}^{\text{CLIP}}(\theta) = \mathbb{E}_t \left[\min\left(r_t(\theta)A_t,\ \text{clip}\quad(r_t(\theta), 1 - \epsilon, 1 + \epsilon)A_t\right)\right],$$

$$\mathcal{L}^{\text{VF}}(\phi) = \mathbb{E}_t \left[\left(V_\phi(s_t) - \widehat{R}_t\right)^2\right],$$

where $r_t(\theta) = \pi_\theta(a_t|s_t)/\pi_{\theta_{\text{old}}}(a_t|s_t)$, $\epsilon = 0.2$, $A_t$ denotes the generalised advantage estimate ($\lambda = 0.95$), and $\widehat{R}_t$ denotes the discounted return.

**Two-stage training pipeline:** Training proceeded in two stages. First, policies were pretrained in a high-fidelity Gazebo simulator (Zhou et al. 2022) using the same robot model and arena layout as in the physical experiments. The simulator was calibrated to the physical platform by system identification, matching friction, motor response and LiDAR noise to minimise the discrepancy between simulated and real trajectories. After calibration, mean absolute error across validation runs was 4.7% for trajectory length and 6.2% for task completion time. Second, pretrained policies were transferred to the physical robots and fine-tuned for 100 episodes of 300 s each. During fine-tuning, on-policy data were collected directly from hardware and network updates were performed every 50 control steps. Unless otherwise stated, the same optimisation settings were used in simulation pretraining and physical fine-tuning.

**Hyperparameters:** Hyperparameters were selected by grid search in simulation on the MH topology with $H = 128$, and were then held fixed across all topologies and hidden sizes. The final settings were: policy learning rate $3 \times 10^{-4}$, value learning rate $1 \times 10^{-3}$, 10 PPO epochs per update, mini-batch size 256, discount factor $\gamma = 0.99$, GAE parameter (Chen et al. 2023) $\lambda = 0.95$, clip range $\epsilon = 0.2$, entropy coefficient 0.01, value-loss coefficient 0.5 and maximum gradient norm 0.5. The policy learning rate used linear warmup (Ma and Yarats 2021; Cheon and Paik 2026) over the first 10% of training steps followed by cosine decay to zero. Optimisation used Adam with $\beta_1 = 0.9$, $\beta_2 = 0.999$ and $\epsilon = 10^{-8}$.

**Training duration:** Simulation pretraining was run for $2 \times 10^6$ environment steps per condition. Physical fine-tuning added 100 episodes for each condition deployed on hardware. Early stopping was applied when the mean evaluation reward over 100 consecutive episodes improved by less than 1%; this occurred only in a subset of the largest simulation-only models. Total training time across all simulated conditions was approximately 3000 GPU hours on two NVIDIA A100-40GB GPUs.

**Random seeds and reproducibility:** A master random seed of 42 was used, and condition-specific seeds were generated deterministically from it. Python, NumPy, PyTorch and Gazebo random number generators were seeded accordingly. Exact seed lists, configuration files and training scripts are provided in the Supplementary Information and code repository.

**Evaluation during training:** During simulation pretraining, policies were evaluated every 50,000 steps on 10 deterministic episodes without exploration noise. During physical fine-tuning, evaluation was performed every 10 episodes on 5 hardware runs. Reported final performance for each condition was computed from the final evaluation window of training. Evaluation seeds were held fixed across matched conditions to ensure comparability across topologies and scales.

## 4.7 Physical experimental protocol

Each physical run followed a standardised protocol designed to minimise uncontrolled variability. Before each run, the 10 robots were placed at random initial positions within a 2 m × 2 m starting zone, with orientations sampled uniformly from $[0, 2\pi)$. The four coloured boxes were placed at random positions within a central 4 m × 4 m region, whereas the four colour-matched goal zones remained fixed at the arena corners. Random initial conditions were controlled by predefined seeds recorded for each run.

After initialisation, the robots were started from a central laptop command and executed the pre-trained policy autonomously for 300 s (1500 control steps at 0.2 s per step), without human intervention. Communication topology was enforced by software routing tables loaded onto each robot's onboard computer. During each run, robots logged pose, selected action, transmitted and received messages, box-state observations and local occupancy-grid snapshots. Logs were synchronised offline after each run.

Ground-truth robot poses were recorded throughout the experiments using an OptiTrack motion-capture system with 12 cameras. Motion-capture data were aligned with onboard logs through timestamps and used to compute map coverage accuracy and verify that odometric drift remained within acceptable limits over the duration of a run.

For the main topology–scale experiment, each combination of topology (4 levels) and hidden size (32, 128 and 512) was evaluated in 5 independent physical runs, giving a total of $4 \times 3 \times 5 = 60$ runs. Runs were distributed across 20 different days to reduce temporal dependence arising from environmental or hardware variation. Random seeds controlling initial robot poses, box positions and goal assignments were recorded for all runs and are provided in the Supplementary Information.

A standard maintenance routine was performed before each day's experiments, including LiDAR alignment, odometry reset and battery checks. These procedures were intended to keep robot behaviour stable across the full experimental period.

The same run protocol was used for robustness tests, with additional packet-loss or dropout perturbations injected as described below.

## 4.8 Composite performance metric

The primary run-level performance metric was a composite score designed to balance task completion, mapping quality and communication efficiency:

$$\text{score} = (\text{transport} \times w_{\text{trans}}) + \text{map} - \text{energy},$$

where transport is the number of successfully delivered boxes (0–4), map is final global occupancy-map coverage accuracy (0–100%), and energy is total communication energy consumed during the run (J). A box was counted as delivered when it remained inside its colour-matched goal zone for at least 5 s. Map coverage accuracy was computed by comparing the final shared occupancy map with a motion-capture-derived ground-truth map. Communication energy was computed by summing the measured per-packet cost over all received messages.

We set the transport coefficient to $w_{\text{trans}} = 10$ to approximately balance the empirical ranges of the three components. Under this weighting, transport, map accuracy and communication energy each contributed on a comparable scale to the raw score, reducing the likelihood that any single term would dominate the comparison.

For reporting, raw scores were linearly rescaled to 0–100 using the observed minimum and maximum over the 60 physical runs:

$$\text{normalised_score} = \frac{\text{score} - \text{score}_{\text{min}}}{\text{score}_{\text{max}} - \text{score}_{\text{min}}} \times 100.$$

The same normalisation constants were applied across all topology and hidden-size conditions.

To assess sensitivity to the transport weighting, we varied $w_{\text{trans}}$ from 1 to 20 and recomputed normalised performance for all 60 physical runs. Across this range, the relative ordering of topologies remained unchanged (MH > SW > ring > FC; Friedman test for rank consistency, $p < 0.001$). We additionally analysed the three sub-metrics separately without weighting. In all cases, MH outperformed FC and ring, indicating that the main conclusion does not depend on the specific composite weighting. Full sensitivity analyses, including bootstrap confidence intervals and sub-metric comparisons, are provided in the Supplementary Information and accompanying code.

## 4.9 Coordination-dynamics measures

To characterise coordination organisation beyond aggregate task performance, we computed two descriptive measures from logged behavioural data: a run-level specialisation index and an inter-module coordination matrix. These measures were used to test whether structured topologies, particularly SW and MH, were associated with more differentiated and more modular coordination patterns.

**Specialisation index:** We quantified behavioural specialisation from each robot's action distribution over a run. For robot $i$, let $f_{i,a}$ denote the frequency of action $a \in \{1, ..., 5\}$ over the episode, and define the corresponding empirical action probabilities as

$$p_{i,a} = \frac{f_{i,a}}{\sum_{a'} f_{i,a'}}.$$

We then defined the robot-level action concentration index as

$$C_i = \sum_{a=1}^{5} p_{i,a}^2 .$$

Higher values of $C_i$ indicate stronger concentration of behaviour onto a smaller subset of actions, and hence stronger within-robot behavioural specialisation. The run-level specialisation index was defined as the mean concentration across the $N = 10$ robots,

$$S = \frac{1}{N} \sum_{i=1}^{N} C_i .$$

Under this definition, $S$ is low when robots use actions relatively uniformly and high when robots show more concentrated behavioural repertoires. We computed $S$ for each physical run and compared topologies using a Kruskal–Wallis test followed by Dunn post-hoc comparisons with Bonferroni correction.

**Inter-module coordination matrix:** For the MH topology, we additionally quantified modular coordination dynamics. Each robot's instantaneous movement was represented by its planar velocity vector $\boldsymbol{v}_i(t) = (v_x(t), v_y(t))$, derived from odometry. For each pair of robots $i$ and $j$, we computed the zero-lag Pearson correlation of their velocity time series over the full episode. Let $M_1$ and $M_2$ denote the two modules of five robots each. We then summarised pairwise correlations within and between modules as

$$C_{pq} = \frac{1}{|M_p||M_q|} \sum_{i \in M_p} \sum_{j \in M_q} \mathrm{corr} \quad (\boldsymbol{v}_i, \boldsymbol{v}_j), \qquad p, q \in \{1,2\}.$$

The resulting $2 \times 2$ symmetric matrix captures the relative strength of intra-module and inter-module coordination. High within-module correlations together with weaker between-module correlations indicate that local subgroups move coherently while remaining only weakly coupled to one another.

Action logs and velocity traces used for these analyses are provided in the repository, together with the scripts used to compute the specialisation index and inter-module coordination matrix. Full implementation details are provided in the Supplementary Information.

## 4.10 Robustness perturbation protocol

To evaluate the resilience of different communication topologies under adverse conditions, we conducted two classes of robustness tests: packet loss and robot dropout. All tests were performed at hidden size 128 using the same trained policies as in the main physical experiment. Each perturbation level was evaluated with 5 independent runs per topology, giving $4 \times 7 \times 5 = 140$ runs for packet loss and $4 \times 6 \times 5 = 120$ runs for dropout. The same randomisation and daily calibration procedures used in the main experiment were retained.

Packet loss was used to model unreliable communication. At each control step, each incoming message was independently dropped before delivery with probability

$$p_{\mathrm{loss}} \in \{0\%, 5\%, 10\%, 15\%, 20\%, 25\%, 30\%\}.$$

Dropped packets were not retransmitted and no acknowledgement mechanism was used. The physical WiFi channel remained active, but messages selected for loss were discarded at the software level. This procedure approximates communication degradation caused by interference, fading or congestion while leaving the underlying communication architecture unchanged.

Robot dropout was used to model partial hardware failure or robot removal. Before the start of each run, a fraction

$$d \in \{0\%, 10\%, 20\%, 30\%, 40\%, 50\%\}$$

of robots was selected uniformly at random without replacement and removed from active participation. Disabled robots produced no actions and transmitted no messages for the duration of the run. The communication graph was then updated over the remaining active robots while preserving the defining principle of each topology. This procedure tests whether the system can maintain useful coordination after partial loss of team members.

For each topology and perturbation type, we defined the empirical collapse point as the smallest perturbation level at which mean normalised performance, averaged over 5 runs, fell below 50% of the corresponding unperturbed baseline for the same topology. When the threshold lay between two sampled perturbation levels, the collapse point was estimated by linear interpolation. These empirical collapse points were compared with theoretical percolation thresholds in the main text. Perturbation schedules, random seeds and implementation details are provided in the Supplementary Information and code repository.

## 4.11 Simulation-calibrated extrapolation

Physical experiments were restricted to hidden sizes $H \leq 512$ because inference time on the Raspberry Pi 4 exceeded 500 ms for $H \geq 1024$, which is incompatible with the 0.2 s control cycle. To explore larger policy networks, we therefore used a high-fidelity Gazebo simulator calibrated to the physical robot platform.

**Calibration and validation:** The simulator was calibrated to real-robot dynamics by system identification using trajectory, velocity and map-coverage data from physical runs. Simulation parameters, including friction, motor response and LiDAR noise, were adjusted to minimise the discrepancy between simulated and real trajectories. After calibration, validation against matched real-world runs yielded mean absolute errors of 4.7% for trajectory length, 6.2% for task completion time and 3.1% for final map coverage. These errors indicate that the simulator captured the principal dynamical features of the physical system sufficiently well to support extrapolative analysis.

**Extrapolation experiment:** For the large-scale extrapolation reported in the main text, we fixed the topology to MH and varied hidden size over $H \in \{32, 128, 512, 1024, 2048, 4096\}$. Each condition was trained with 5 random seeds for $2 \times 10^6$ environment steps using the same optimisation settings as in the physical-task training pipeline. Environment randomisation was matched as closely as possible to the physical experiments.

**Interpretation:** All results for $H \geq 1024$ are interpreted as *simulation-supported extrapolation* rather than direct physical evidence. Although calibration error was low, unmodelled real-world effects, including thermal throttling, communication interference and sensor drift, could shift the point at which scaling gains saturate.

Accordingly, the reported saturation beyond 1024 hidden units should be interpreted as simulation-based evidence for diminishing returns under the tested task and architecture, rather than as a direct hardware-level demonstration. Calibrated simulation parameters, validation data and training logs are provided in the repository and Supplementary Information.

## 4.12 SMAC replication experiments

To test whether the main qualitative pattern reproduced beyond the physical robot task, we performed independent experiments on two StarCraft Multi-Agent Challenge (SMAC) maps: `3m_vs_3m` and `2s_vs_1sc` (Vinyals et al. 2019). These maps were selected to provide a standard multi-agent reinforcement-learning setting with partial observability and cooperative control, while differing in symmetry and tactical demands.

All experiments used QMIX (Rashid et al. 2020) with matched hyperparameters across conditions. Each topology–map condition was trained for $2 \times 10^6$ environment steps with 5 random seeds. Test win rate against the built-in game AI was used as the primary evaluation metric. To compare topology effects with scale effects, we additionally increased the FC hidden size from 64 to 256 under otherwise matched settings.

For topology comparisons, we considered three interaction structures: fully connected (FC), ring and modular hierarchical (MH). Topologies were implemented by masking inter-agent communication according to fixed adjacency matrices. In `3m_vs_3m`, all three allied Marines were controlled agents, allowing a direct comparison among FC, ring and MH. In `2s_vs_1sc`, only two allied Stalkers were controlled, so the scope for distinct topology structure was reduced; accordingly, this map should be interpreted as a lower-dimensional replication setting rather than as a topology contrast of equal structural richness to `3m_vs_3m`. Where MH was used, agents were grouped by unit role whenever such grouping was meaningful.

Evaluation was performed periodically during training, and final performance was summarised from the terminal evaluation phase of each run. These SMAC experiments were treated as independent replications of the physical-task result rather than as part of the primary controlled experimental pipeline. Implementation details, configuration files and random seed lists are provided in the code repository and Supplementary Information.

## 4.13 Benchmark reanalysis procedure

As secondary supporting consistency checks, we reanalysed three publicly available multi-agent benchmarks spanning distinct task domains, observation spaces and evaluation protocols. These benchmarks were not part of the primary controlled experimental pipeline. Their purpose was to test whether the broad qualitative contrast between interaction structure and agent scale remained consistent across heterogeneous settings.

We selected three benchmarks on the basis of data availability and task diversity: Overcooked (Carroll et al. 2019), Google Research Football (Kurach et al. 2020), and Kilobot swarm assembly (Rubenstein et al. 2014). For each benchmark, we identified one comparison representing a structural change in inter-agent interaction and one comparison representing an increase in agent scale under the closest available matched setting. Performance was evaluated using the benchmark-specific primary outcome reported in the source study.

For each benchmark, we extracted mean performance values together with corresponding measures of uncertainty from figures or tables in the original publications. Where raw numerical values were not directly available, data were digitised from published figures. We then computed Cohen's $d$ for the structure comparison and for the scale comparison as

$$d = \frac{\mu_2 - \mu_1}{\sigma_{\text{pooled}}}, \qquad \sigma_{\text{pooled}} = \sqrt{\frac{(n_1 - 1)s_1^2 + (n_2 - 1)s_2^2}{n_1 + n_2 - 2}},$$

where $\mu_1, \mu_2$ are the group means, $s_1, s_2$ are the group standard deviations and $n_1, n_2$ are the corresponding sample sizes. When studies reported standard errors rather than standard deviations, these were converted using $s = \text{SE}\sqrt{n}$.

Because the selected benchmarks differ substantially in task structure, evaluation protocol and sample size, we synthesised effect sizes using a random-effects meta-analysis. This model was used to summarise the average contrast between structure changes and scale increases while allowing true effects to vary across benchmarks. Heterogeneity was quantified with $I^2$, and the resulting benchmark synthesis is interpreted as a consistency check rather than as decisive evidence. Digitised data, extracted summary statistics and meta-analysis code are provided in the repository and Supplementary Information.

## 4.14 Statistical analysis

The primary analysis unit was the run-level aggregate performance score. For the physical topology–scale experiments, we fitted linear mixed models using topology as a categorical fixed effect, $\log_2$(hidden size) as a continuous fixed effect, and day of experiment as a random intercept to account for day-to-day variability. Nested models were compared by likelihood-ratio tests, and marginal $R^2$ values were used to summarise the relative explanatory contributions of topology and scale.

Non-parametric analyses were used for coordination-dynamics measures, robustness comparisons and sub-metric sensitivity analyses. Kruskal–Wallis tests were followed by Dunn post-hoc comparisons with Bonferroni correction where appropriate. Rank consistency across alternative transport weights in the composite metric was assessed using Friedman tests. Piecewise regression was used for the simulation-based saturation analysis, and benchmark-level effect sizes were synthesised with a random-effects meta-analysis.

All tests were two-tailed with $\alpha = 0.05$, with multiple-comparison correction applied where indicated. Primary controlled evidence was taken from the physical robot experiments. Simulation results were interpreted as extrapolative support beyond the hardware-feasible regime, SMAC analyses as independent replications, and public benchmark reanalyses as supporting consistency checks only.

# 5 Conclusion

In a real-world multi-robot system, we find that structured interactions can contribute more to distributed coordination than increasing individual model scale under matched hardware budgets. Across controlled physical experiments, modular hierarchical communication consistently outperformed fully connected and ring topologies, whereas gains from enlarging onboard neural networks were comparatively modest. This qualitative pattern remained robust across metric definitions and robustness perturbations, and was reproduced in independent SMAC experiments.

These results identify interaction structure as a consequential system-level design axis for multi-robot coordination in the tested setting. At the same time, the quantitative scope of our conclusions remains bounded by the specific system instance, task family and hardware regime studied here, and the apparent saturation of scaling gains at larger model sizes is supported by simulation rather than direct physical deployment. More broadly, our findings suggest that, in resource-constrained multi-robot settings, redesigning how robots interact may be a more effective first intervention than further scaling individual policies.

## 6 Acknowledgements

The authors would like to express their sincere gratitude to the anonymous reviewers for their insightful comments and constructive feedback, which greatly contributed to improving the overall quality of this paper. This work was supported in part by the National Key Research and Development Program of China under Grant 2022YFF0903300; and in part by National Natural Science Foundation of China under Grant 92167109.

## Author contributions

J.W. was responsible for conceptualisation, methodology, project administration, supervision and writing of the original draft. Y.T., G.Z. and J.Z. contributed equally to investigation, experimental execution, software implementation, data collection and data preprocessing. G.Z. and J.Z. contributed to model implementation, training and evaluation pipelines, multi-agent learning baselines, simulation experiments and policy evaluation. Y.T. contributed to multi-robot system integration, physical-robot deployment, sensor calibration, motion-capture processing and physical-experiment validation. Y.L. contributed to data curation, simulation assets, benchmark organisation, repository preparation and reproducibility checks. S.S. contributed to conceptualisation, methodology, supervision and interpretation of results. Y.M. contributed to conceptualisation, methodology, supervision, project coordination and interpretation of results. J.W., G.Z., J.Z. and Y.L. performed formal analysis, benchmark reanalysis and visualisation. J.W., S.S. and Y.M. jointly supervised the study and served as corresponding authors. All authors contributed to reviewing and editing the manuscript and approved the final version.

## Competing interests

The authors declare no competing interests.

# Data and Code Availability

The data and code supporting this study are available during peer review through an anonymised OSF view-only repository:

The repository, entitled *Distributed Intelligence_repository*, is organised as a reproducibility package and contains physical-run data, motion-capture split archives, analysis scripts, training and evaluation code, figure data and regeneration scripts, simulation world files, URDF and mesh assets, SMAC supporting benchmark data and configuration files, documentation, manifests and validation reports. The manuscript source files are not included in the repository. An executable Code Ocean capsule package is also provided in the repository for reviewer-side reproduction of the statistical analyses, benchmark reanalyses and figure-generation workflows. Full physical-robot execution, complete ROS/Gazebo deployment and full-scale training require external hardware/runtime dependencies and are documented separately. Upon acceptance, the repository will be made publicly available and archived with a persistent DOI.